%% file: main.tex
\documentclass{ieeeaccess}

\usepackage{natbib}
\usepackage{amsmath,amssymb,amsfonts}
\usepackage{graphicx}
\usepackage{colortbl}

\usepackage{algorithm}
\usepackage{algpseudocode}
\usepackage{lscape}
\usepackage[inline]{enumitem}
\usepackage[utf8]{inputenc}
\usepackage{booktabs}
\usepackage{multirow}
\usepackage{array}
\usepackage[T1]{fontenc}    
\usepackage{url}            
\usepackage{nicefrac}       

\usepackage{caption}
\newcolumntype{C}[1]{>{\centering \arraybackslash}m{#1\textwidth}}
\newcolumntype{R}[1]{>{\raggedleft \arraybackslash}m{#1\textwidth}}
\newcolumntype{K}[1]{>{\centering \arraybackslash}m{#1\columnwidth}}
\newcolumntype{L}[1]{m{#1\columnwidth}}

\usepackage{textcomp}
\def\BibTeX{{\rm B\kern-.05em{\sc i\kern-.025em b}\kern-.08em
    T\kern-.1667em\lower.7ex\hbox{E}\kern-.125emX}}
    
\usepackage{hyperref}       

\begin{document}

\history{Date of publication xxxx 00, 0000, date of current version xxxx 00, 0000.}
\doi{10.1109/ACCESS.2017.DOI}
\title{Learning the Quality of Machine Permutations in Job Shop Scheduling}
\author{\uppercase{Andrea Corsini}\authorrefmark{1},
\uppercase{Simone Calderara\authorrefmark{2}}\IEEEmembership{Member, IEEE}, and \uppercase{Mauro Dell'Amico}\authorrefmark{3}}
\address[1]{University of Modena and Reggio Emilia (e-mail: andrea.corsini@unimore.it) 
}
\address[2]{University of Modena and Reggio Emilia (e-mail: simone.calderara@unimore.it) 
}
\address[3]{University of Modena and Reggio Emilia (e-mail: mauro.dellamico@unimore.it)
}
\tfootnote{This paragraph of the first footnote will contain support 
information, including sponsor and financial support acknowledgment. For 
example, ``This work was supported in part by the U.S. Department of 
Commerce under Grant BS123456.''}

\markboth
{Author \headeretal: Preparation of Papers for IEEE TRANSACTIONS and JOURNALS}
{Author \headeretal: Preparation of Papers for IEEE TRANSACTIONS and JOURNALS}

\corresp{Corresponding author: Andrea Corsini (e-mail: andrea.corsini@unimore.it).}

\input{sections/0_abstract}


\titlepgskip=-15pt

\maketitle

\input{sections/1_intro}

\input{sections/2_related}
\input{sections/3_proposal}
\input{sections/4_results}
\input{sections/5_close}

\appendices
\input{sections/AppA}
\input{sections/AppB}
\input{sections/AppC}

\bibliographystyle{unsrtnat}
\bibliography{main}

\begin{IEEEbiography}[{\includegraphics[width=1in,height=1.25in,clip,keepaspectratio]{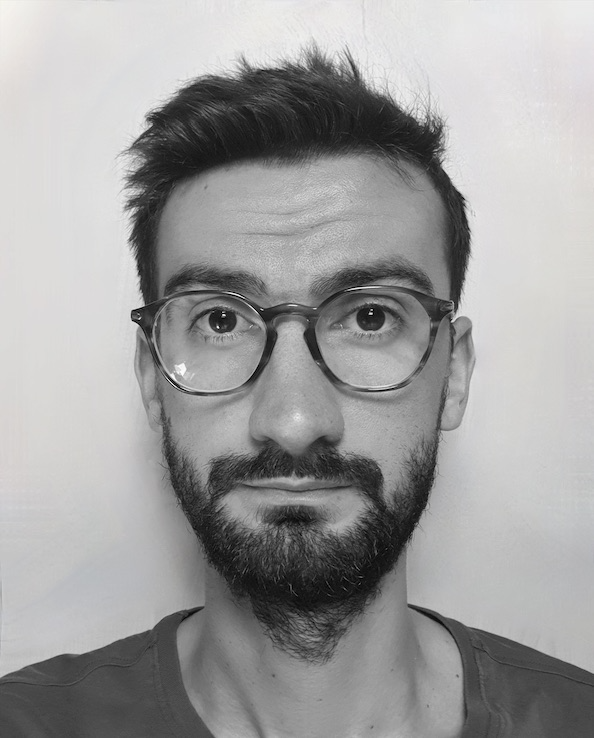}}]{Andrea Corsini} received the B.S. (2018) and M.S. (2020) in computer engineering from the University of Modena and Reggio Emilia, Modena, Italy. 
He is now pursuing his Ph.D. in industrial innovation engineering at the University of Modena and Reggio Emilia.
His current research interests include operations research and machine learning, with a particular focus on how to apply deep learning for solving combinatorial optimization problems.
\end{IEEEbiography}

\begin{IEEEbiography}[{\includegraphics[width=1in,height=1.25in,clip,keepaspectratio]{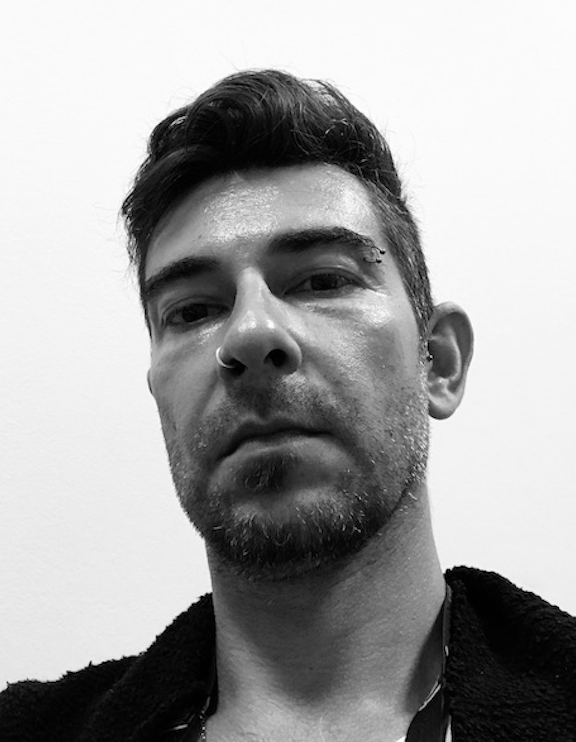}}]{Simone Calderara} (Member, IEEE) received the master’s degree in computer engineering and the PhD degree from the University of Modena and Reggio Emilia, Modena, Italy, in 2005 and 2009, where he is currently an assistant professor within the Imagelab Group. 
His current research interests include computer vision and machine learning applied to human behavior analysis, visual tracking in crowded scenarios, and time series analysis for forensic applications.
\end{IEEEbiography}

\begin{IEEEbiography}[{\includegraphics[width=1in,height=1.25in,clip,keepaspectratio]{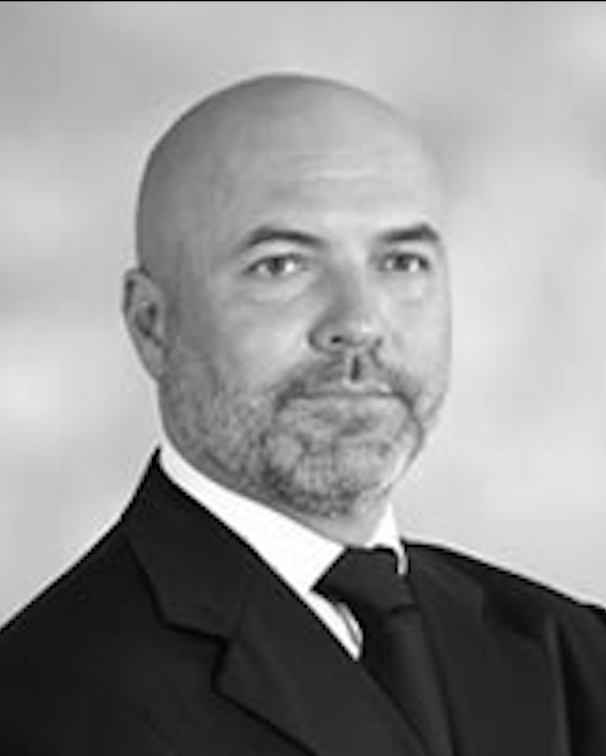}}]{Mauro Dell'Amico} is currently a Full Professor of operational research with the University of Modena and Reggio Emilia. 
He has almost three decades of academic experience in combinatorial optimization and operations research, primarily applied to mobility, logistics, transportation, supply chain management, production scheduling and planning, and network planning. 
He has participated as a principal investigator in many EU and Italian funded research projects in optimization, logistics, ICT, transportation, and scheduling. 
He combines the academic activities with consultancy on optimization for private and public companies. He is a member of the scientific board of several conferences and journals in operations research.
\end{IEEEbiography}

\EOD

\end{document}

%% file: sections/0_abstract.tex
\begin{abstract}
In recent years, the power demonstrated by Machine Learning (ML) has increasingly attracted the interest of the optimization community that is starting to leverage ML for enhancing and automating the design of algorithms.
One combinatorial optimization problem recently tackled with ML is the Job Shop scheduling Problem (JSP).
Most of the works on the JSP using ML focus on Deep Reinforcement Learning (DRL), and only a few of them leverage supervised learning techniques.
The recurrent reasons for avoiding supervised learning seem to be the difficulty in casting the right learning task, i.e., what is meaningful to predict, and how to obtain labels.
Therefore, we first propose a novel supervised learning task that aims at predicting the quality of machine permutations.
Then, we design an original methodology to estimate this quality, and we use these estimations to create an accurate sequential deep learning model (binary accuracy above 95\%).
Finally, we empirically demonstrate the value of predicting the quality of machine permutations by enhancing the performance of a simple Tabu Search algorithm inspired by the works in the literature.
\end{abstract}

%% file: sections/1_intro.tex
\section{Introduction}\label{sec:intro}

Nowadays, manufacturing and service industries are becoming larger, more interconnected, and generate every day a large volume of data.
This increase in industrial complexity and the shift towards a 4.0 environment pose new challenges in scheduling and demands new personalized algorithms to maximize production while minimizing costs and processing times \citep{JSP4.0}.

In recent years, there has been a surge of new techniques that take advantage of data generated by smart devices, sensors, and industrial systems.
The discipline encompassing much of these techniques is machine learning.
Machine learning demonstrated how data can be fruitfully used to achieve astonishing results in fields like computer vision and natural language processing.
Based on this premise, ML constitutes a concrete opportunity to answer the new industrial demands.

However, ML is not yet mature and ubiquitous in all fields.
One of these fields is combinatorial optimization, where only recent works achieve superior performance compared to few non-ML algorithms in problems like the travelling salesman problem \citep{TSP,GNNDRL}, the vehicle routing problem \citep{VRP}, scheduling \citep{AC-DRL,DispatchDRL,ScheduleDRL}, and others \citep{Horizon,RLCOSurvey}.
These pioneering works demonstrated how ML can be applied to combinatorial problems, but, due to the limitations of these works and the partial coverage of the many ML paradigms, much more has to be discovered.

In this work, we focus on the \emph{Job Shop scheduling Problem}~\citep{Scheduling}, a notorious NP-hard combinatorial problem with many practical applications in industry.
Simply put, the JSP is to schedule a set of jobs onto a set of machines by minimizing an objective function.
The distinctive characteristic of the JSP is that each job consists of a strict chain of operations, each of which has to be processed on one and only one machine without interruptions (see Section~\ref{ssec:JSP} for the formal definition).

Mixed Integer Linear Programming (MILP) and Constraint Programming are two exact optimization methods to solve the JSP \citep{MIP}.
Although these methods are becoming everyday faster, they do not scale well on medium and large instances \citep{MIP}, and they become rapidly useless even in small but complex industrial environments \citep{JSP4.0}.
For these reasons, approximation methods are still largely employed and constitute an active area of research, besides being one of the subjects of this work.

Most of the recent ML-based works tackling the JSP rely on Deep Reinforcement Learning (DRL) techniques \citep{AC-DRL,DispatchDRL,ScheduleDRL}. 
What makes (deep) reinforcement learning particularly appealing in the context of the JSP is its ability to learn from past decisions, without the need of labels and by correctly formulating the Markov Decision Process~\citep{Sutton}.
However, training effective DRL agents is a difficult optimization task, it is not easy to reproduce \citep{DRLRepro}, and takes a lot of time \citep{DispatchDRL}, especially for Monte Carlo-based methods \citep{Sutton}.
Therefore, we investigate herein whether it is possible to use a supervised learning approach to solve the JSP.

Our work has been guided by two fundamental questions: \begin{enumerate*}[label=(\roman*)]
    \item what type of information might be used or might help solve a JSP instance?
    \item is it possible to learn this information in a supervised manner?
\end{enumerate*}
These questions arise from the fact that not all the solutions to a JSP instance are feasible, i.e., respect the problem constraints, and for those feasible, the objective value (e.g., the makespan or the total tardiness) is not trivially derivable.
For these reasons, the application of supervised learning to the JSP requires a learning task tightly related to the objective function, and that may fit in the back-propagation algorithm.
We thus propose as a novel supervised learning task to \emph{learn the quality of a machine permutation, i.e., how good is the sequence of operations on a machine}.

Understanding whether a sequence of operations on a machine is of ``high quality'' is a difficult task in the JSP \citep{Scheduling}.
Since having a method to judge machines is important, either for speeding up existing algorithms or even in machine-based decomposition, we present an original methodology to learn the quality of machines by means of \emph{sequential deep learning} and a \emph{MILP solver}.
There already exists in literature approaches to evaluate a machine, most notably~\citep{bottleneck}, but they frequently estimate the criticality of machines, a related but different concept.
Contrary, we define the quality of a machine permutation as the \emph{likelihood of finding this permutation in an optimal or near-optimal solution}.

We evaluate the impact of our proposal by comparing the results obtained with one of the best metaheuristics for the JSP, namely the Tabu Search (TS), with and without these quality estimations.
In addition, we compare the results of the TS with some of the DRL approaches to justify our proposal for enhancing existing approximation algorithms.

Summarizing, the contributions of this work are:
\begin{enumerate*}[label=(\roman*)]
    \item we propose a novel supervised learning task for the JSP;
    \item we propose an original methodology to evaluate the quality of machine permutations by means of a MILP solver;
    \item we create a supervised dataset on which we train a sequential deep learning model;
    \item we test the advantages of our learning task in a Tabu Search algorithm.
\end{enumerate*}
In the remainder of this work, we start by describing in Section~\ref{sec:related} some of the start-of-the-art algorithms to solve the JSP and recent trends leveraging ML.
In Section \ref{sec:proposal}, we present the mathematical intuition behind our learning task, the methodology to estimate the quality of machine permutations, the sequential deep learning model, and the Tabu Search.
Finally, in Section~\ref{sec:results} we present the dataset, the performance of the learning model, the advantages of our proposal when used in a TS, and the comparison with the DRL approaches. In Section~\ref{sec:close} we conclude with few considerations and future works.

\subsection{Job Shop formulation}\label{ssec:JSP}

The Job Shop Problem is as follows: we are given a set of $n$ jobs $J = \{ 1, \dots, n \}$, and a set of $m$ machines $M = \{1, \dots, m \}$.
Each job $j \in J$ is composed by a sequence of $m_j \in \mathbb{N}$ operations $O_j = ( o_j^1, \dots, o_j^{m_j} )$ that specifies in which order the jobs must be processed on the machines.
Thus, an operation $o_j^i$ belongs to job $j$, needs to be processed on machine $\mu_j^i \in M$ and has processing time $p_j^i \in \mathbb{R}_{>0}$.
In this work, we consider as the objective of the JSP the \emph{minimization of the makespan}, that is the total length in time required to complete all the jobs. 
Preemption is not allowed, and machines can handle one operation at a time.
In scheduling theory, this problem is identified as $Jm \, || \, C_{max}$.

Solving a JSP instance means finding a permutation of operations on each machine such that the makespan is minimized, the precedence among the operations are respected, and the operations do not overlap on each machine.
Let $\Pi = \{ \pi_1, \dots, \pi_m \}$ be a solution of a JSP instance, and $\pi_i = (s_1^i, s_2^i, \dots, s_{n_i}^i)$ the permutation or sequence of the $n_i \in \mathbb{N}$ operations on machine $i \in M$.
The permutation $\pi_i$ fixes the order of operations on machine $i$, and $s_k^i$, with $k \in \{1, \dots n_i \}$, gives the operation of some job $j$ that is processed in the $k^{th}$ position.

\Figure[t!](topskip=0pt, botskip=0pt, midskip=0pt)[width=0.99\columnwidth]{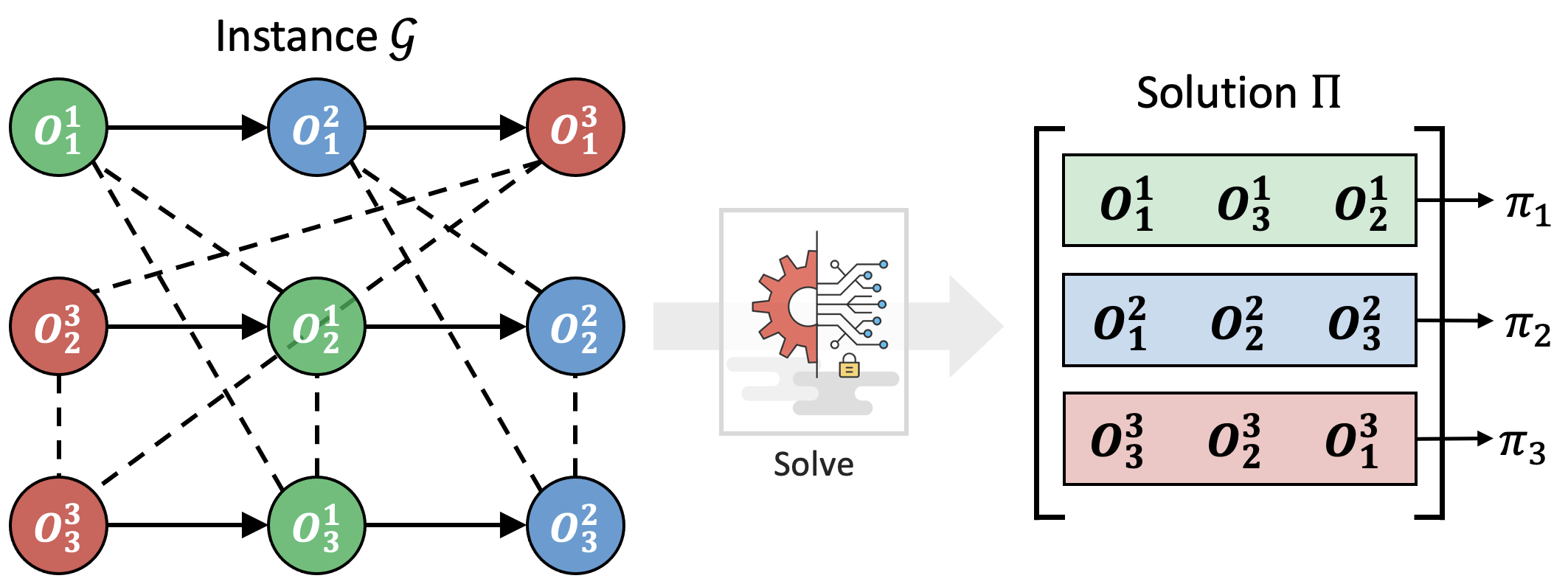}
{On the left, an example of a disjunctive graph that represents a JSP instance with 3 jobs and 3 machines. On the right, a feasible solution that gives the sequence of operations for each machine.
\label{fig:graph}}

It is common to represent a JSP instance as a \emph{disjunctive graph} $\mathcal{G} = (V, A, E)$ (see Figure~\ref{fig:graph}); where $V$ is the set of operations, $A$ is the set of arcs connecting consecutive operations of the same job, and $E$ is the set of disjunctive edges connecting operations to be processed on the same machine.
In this representation, the problem of minimizing the makespan is reduced to finding an orientation to the edges in $E$ such that the weighted longest path (a.k.a. the \emph{critical path}) is minimized, where weights are the processing times of operations.
We will refer to this set of oriented edges with $\hat{E}$ and the corresponding digraph with $\hat{\mathcal{G}}=(V, A, \hat{E})$.
Finally, note there is a unique one-to-one correspondence between a generic solution $\Pi_g$ and a digraph $\hat{\mathcal{G}}_g$ (orienting the edges of $E$ is equivalent to creating permutations $\pi_i$ and vice versa), and if this digraph is acyclic the solution is guaranteed to be feasible \citep{Scheduling}.

%% file: sections/2_related.tex
\section{Related Works}\label{sec:related}

In this section, we review two popular approximation methods for the JSP, namely \emph{Priority Dispatching Rules} (PDR) and \emph{Metaheuristics}, and some recent trends that leverage machine learning in such methods.

\subsection{Priority Dispatching Rules}\label{ssec:PDR}

A priority dispatching rule~\citep{PDR} is a heuristic method that assigns operations to machines based on priorities.
In general, priorities are assigned with hard-coded rules that consider the status of the schedule or characteristics about jobs, machines, and operations.
Designing an effective PDR is difficult and requires substantial domain knowledge, especially on complex problems like the JSP.
Moreover, the performance of a PDR often varies drastically in different instances.
Therefore, in the last decade, many researchers tried to automate the design of PDRs with the help of machine learning.

One of the first applications of ML to PDRs is presented in \citep{PDRNN}, where a neural network selects the most suited PDR among a pool of rules.
The decisions of the neural network are based on the current system state and the training phase is done through simulations.
In \citep{ImitationPDR}, an imitation learning method is proposed to learn PDRs by using the supervision of a MILP solver.
This work demonstrated how learning from optimal solutions is not enough to produce robust PDRs.

Most of the recent research efforts focus on adapting DRL to learn PDRs.
After a correct formulation of the Markov Decision Process, a policy to schedule operations is learnt from the experience derived by resolving the same instances many times.
In \citep{AC-DRL}, an actor-critic architecture \citep{ActorCritic} is proposed, where the critic evaluates the value of decisions in the partial schedule, whereas the actor learns to make decisions based on the schedule and the critic estimations.
In \citep{DispatchDRL}, an actor-critic is also proposed, but with a Graph Neural Network \citep{GNN} (GNN) to construct an adaptive representation of the partial schedule.
One interesting aspect of this work is that the authors underline how GNNs seem to have poor performance when applied to disjunctive graphs.
Another example of an actor-critic architecture coupled with a GNN is in \citep{ScheduleDRL}.
This work applies GNNs to the disjunctive graphs of JSP instances by specifically designing a GNN architecture and by using a rich set of features to describe operations.
From these works, it is not possible to draw any conclusion on the benefits of applying GNNs to disjunctive graphs, therefore, we prefer to avoid GNNs.

Although these promising works showed how to create superior PDRs, the performance of these proposals is still much worse than the performance of metaheuristics.
Due to their lower performance and the lack of guarantees of producing high-quality solutions, PDRs still remain a valid alternative as generators of initial solutions for metaheuristics.

\subsection{Metaheuristics}\label{ssec:LSA}

The general idea of a metaheuristic~\citep{Metaheuristic,LSA} is to describe trajectories in the solution space starting from initial solutions and visiting neighbor solutions according to some criteria.
Each trajectory generally stops either when no improving solution exists in the neighborhood, i.e., the current solution is a local optimum, or when a predefined criterion is met.
The effectiveness of metaheuristics depends on a brittle and complex balance of its elements that governs the creation of successful trajectories.
This balance is achieved by designing elements like the neighborhood structure, the searching procedure, and other mechanisms such that the algorithm can intensify promising regions while escaping from local optima.
Therefore, selecting, designing, and assembling the right elements is extremely important and requires domain and algorithm-design expertise.

The breakthrough work in the field of metaheuristics for the JSP is \citep{SA}.
This work adapted the Simulated Annealing (SA)~\citep{firstSA} and proposed one of the most studied and effective neighborhood structures for the JSP, called N1.
N1 was the first to show how it is possible to construct the neighborhood of a solution without incurring in unfeasible solutions.
In addition, it guarantees the existence of a trajectory that leads to global minima, the so-called \emph{convergence property}.

After this work, many variations and extensions of N1 were proposed in \citep{TABU,TSAB,N7}, mostly in the context of a Tabu Search~\citep{TS}.
The most successful application of the TS to the JSP is \citep{TSAB}, where the authors proposed a reduced variation of N1 in which some of the neighbor solutions were removed since they cannot immediately improve the current solution.
In \citep{TSAB} is also proposed the best implementation of the TS for the JSP, successively refined in \citep{iTSAB} by incorporating elements of path relinking in the generator of initial solutions.

Besides the TS and SA, there are other metaheuristics proposed to tackle the JSP \citep{AATS,PSOTS,SurveyGA}.
In these regards, we just want to stress that regardless of the metaheuristics, e.g., Single-Source or Population-Based~\citep{Metaheuristic}, an ad-hoc searching procedure or a local search is often required to enhance performance \citep{AATS,PSOTS,SurveyGA}.

As reviewed in \citep{MLintoHeuristics}, ML can be fruitfully integrated in the most common metaheuristics and constitutes an opportunity to enhance, simplify, and automate the creation of effective algorithms.
Some examples of how ML techniques can be integrated into metaheuristics for scheduling problems are \citep{LocalRewriting,LearningVNS}.

In \citep{LocalRewriting}, it is proposed a DRL-based rewriting method in which a region-picking policy selects regions of solutions that are rewritten with rules selected by a rule-picking policy.
Picking the right regions and selecting the best rewriting rule are non-trivial operations, and learning to perform them from experience outperformed heuristic rules.
In \citep{LearningVNS}, a Variable Neighborhood Search is enhanced with a mechanism that favors the creation of solutions having promising attributes during the shaking step.
Although this work does not use any ML techniques, learning to construct these solutions might be a viable and better approach.
For other examples of how to combine ML with metaheuristics, we refer the reader to \citep{MLintoHeuristics}.

Despite these premises, metaheuristics did not receive the same attention as PDRs in hybridization with ML for the JSP, and we believe there is much to gain from such a combination.

%% file: sections/3_proposal.tex
\section{Proposed methodology}\label{sec:proposal}

This section starts by outlining the proposed learning task and the mathematical intuition behind it.
Then, we describe our methodology to evaluate the quality of machine permutations and the learning model to tackle the proposed task.
Finally, we present the TS algorithm used to validate the advantages brought by our learning task.

\subsection{Learning Task}\label{ssec:task}

Our novel supervised learning task about the JSP is to \emph{predict the quality of machine permutations, where the quality of a permutation is the likelihood of finding this permutation in an optimal solution}.
We arrived at this formulation after carefully reviewing the abundant literature about the JSP in search of an answer to the first question of Section~\ref{sec:intro}: what type of information might be used for solving the JSP.
To justify why our learning task should help in solving the JSP, 
we briefly report the intuition behind the proof of the convergence property of the N1 neighborhood (see \citep{SA} for the complete proof).

Let $\Pi_1$ and $\Pi_o$ be respectively a feasible solution and an optimal solution of an instance.
The converge property implies that from any $\Pi_1$, it is possible to construct a trajectory of solutions through N1 that allows moving from $\Pi_1$ to an optimal solution $\Pi_o$.
The proof starts from the definition of a special set of critical arcs (remember that critical arcs are those arcs on the longest path in $\hat{\mathcal{G}}$):\par
{\small
\begin{equation}
    K_1(\Pi_o) = \{ (v, w) \in \hat{E_1}  \, | \, (v, w) \text{ is critical} \, \wedge \, (w, v) \in \hat{E_o} \}
\end{equation}
}
that is the set of critical arcs in $\mathcal{\hat{G}}_1$ that do not belong to the optimal solution $\mathcal{\hat{G}}_o$.
When $\Pi_1 \neq \Pi_o$, this set is always non-empty, and it is possible to create a finite trajectory $(\Pi_1, \Pi_2, \dots, \Pi_o)$ that guarantees to reach an optimal solution, where $\Pi_2$ is obtained from $\Pi_1$ by reversing an arc in $K_1$.
Clearly, the convergence is a desirable property for a neighborhood structure, but in practice, it is of no help because it requires to know the set of critical arcs to reverse, i.e., it requires $K$.

Nonetheless, this proof leads us to what might be beneficial for solving the JSP: \emph{an information about which critical arcs are unlikely to be in an optimal solution}. 
At least in the context of N1, knowing this information allows excluding those solutions that introduce arcs unlikely to be in $\hat{E_o}$, resulting in better exploration and a faster convergence towards optima.
However, there is a problem in learning a function that gives the likelihood of finding an arc in an optimal solution: the representation of the arc must encode enough information about the entire solution.

Instead of learning this function, we propose to learn a function that receives in input the machine permutation associated with an arc and outputs the likelihood of finding this permutation in an optimal solution. 
If a machine permutation resulting from the inversion of a critical arc is of higher quality than the original permutation, the reversed arc has a higher chance of being in $\hat{E_o}$.
Therefore, learning such a function still allows to discriminate which critical arcs should be reversed.
In addition, it simplifies the learning task since a permutation intrinsically encodes more information about the entire solution than a single arc.

Based on this theoretical intuition, our learning task should help solve the JSP in at least those approximation algorithms based on N1.
Note that the proposed learning task might also benefit other approximation methods, for instance, machine-based decomposition and ruin-and-recreate algorithms~\citep{RR}, but proving this is outside the scope of this work.

\subsection{Quality of Machine Permutations}\label{ssec:methodology}

Up to this point, we presented our novel learning task, and we justified why this task should help solve the JSP.
What remains uncovered is how the quality $y_k$ of a machine permutation $\pi_k$ can be quantified.
To define the quality $y_k$, we rely on the concept of makespan, and we compute:\par
{\small 
\begin{equation}
    y_k = 1 - \tanh\left( \frac{C_{max}(\pi_k)}{C_{max}^{\,opt}} - 1 \right)
    \label{eq:label}
\end{equation}\par
} %
where $C_{max}(\pi_k)$ is the best makespan found by imposing $\pi_k$ as part of the solution, $C_{max}^{\,opt}$ is the optimal makespan of the instance, and $\tanh$ is the hyperbolic tangent function.

Note that Equation~\ref{eq:label}, beyond giving the mathematical definition of the quality of a machine permutation, also points out the methodology needed to estimate this quality.
This methodology includes a method to optimally solve the JSP and a method to find the best solution with an imposed sequence $\pi_k$. 
With these methods, Equation~\ref{eq:label} estimates the quality $y_k$ by comparing the best makespan found with the sequence $\pi_k$ against the optimal makespan, and it scales this comparison with the $tanh$ function.
When $\pi_k$ is near-optimal, meaning that $C_{max}(\pi_k)$ is close to the optimal makespan, $y_k$ takes a value close to 1.
Contrary, when $C_{max}(\pi_k)$ is far from the optimal value, $y_k$ takes a value close to 0.
Due to its definition, the quality of a permutation is always a value in the interval $[0, 1] \subset \mathbb{R}$, thus, it can be interpreted as a kind of probability (or a likelihood parameterized by some parameters) of finding the permutation in an optimal solution.

As the method to optimally solve the JSP, we propose to use a MILP solver by formulating the problem as a disjunctive model \citep{MIP}.
As pointed out in \citep{MIP}, today solvers can solve instances with 10 jobs and 10 machines in few seconds.

Instead, as the method to find the best makespan $C_{max}(\pi_k)$ by imposing a sequence $\pi_k$, we propose to use a modified version of the standard disjunctive model, again in a MILP solver. 
In this modified version, we introduce a set of constraints to prevent the solver from changing the order of the sequence $\pi_k$.
Note that this modification effectively reduces the solution space and speeds up the solver.
The modified disjunctive model is then:

{\footnotesize
\begin{align}
    &\text{min} && C_{max}(\pi_k) \label{eq:MP1} \\ 
    & \text{s.t.} && x_j^{o_j^h} \geq x_j^{o_j^{h-1}} + p_j^{o_j^{h-1}} && \forall j \in J, h=2, \dots, m_j \label{eq:MP2} \\
    &&& x_j^i \geq x_k^i + p_k^i - Q \, z_{jk}^i && \forall j,k \in J, j < k, i \in M \label{eq:MP3} \\
    &&& x_k^i \geq x_j^i + p_j^i - Q \, (1 - z_{jk}^i) && \forall j,k \in J, j < k, i \in M \label{eq:MP4} \\
    &&& x_{s_h^i}^i \geq x_{s_{h-1}^i}^i && i \in M, h=2, \dots, n_i \label{eq:MP7} \\
    &&& C_{max}(\pi_k) \geq x_j^{o_j^{m_j}} + p_j^{o_j^{m_j}} && \forall j \in J \label{eq:MP6} \\
    &&& z_{jk}^i \in \{0, 1\} && \forall j,k \in J, i \in M \label{eq:MP8} \\
    &&& x_j^i \geq 0 && \forall j \in J, i \in M \label{eq:MP9}
\end{align}
}
The model has two decision variables: $x_j^i$ gives the starting time of job $j$ on machine $i$, and, $z_{jk}^i$ takes value 1 if job $j$ precedes job $k$ on machine $i$.
The set of constraints (\ref{eq:MP2}) guarantees that for each job, the start time of every operation must be equal to or higher than the completion time of the previous operation.
The disjunctive constraints in sets (\ref{eq:MP3}) and (\ref{eq:MP4}) guarantee that the start time of an operation $o_j^i$ must be higher than the completion time of another operation $o_k^i$ when $o_j^i$ is scheduled before $o_k^i$ and vice versa.
Finally, the set of constraints (\ref{eq:MP7}) fixes the order of operations on machine $i$ to be equal to $\pi_k = (s_1^i, s_2^i, \dots, s_{n_i}^i)$, and the set (\ref{eq:MP6}) computes the makespan.
The value of $Q$ is set to $\sum_{j \in J} \sum_{i \in M} p_j^i$ to ensure the correctness of the disjunctive constraints.

Summarizing, the methodology to obtain the quality of a machine permutation $\pi_k$ starts by optimally solving the JSP instance, then the best makespan $C_{max}(\pi_k)$ is found with the presented modified disjunctive model, and finally, the quality is computed with Equation~\ref{eq:label}.

\subsection{The Learning Model}\label{ssec:oracle}

In order to predict the quality $y_k$ of a sequence $\pi_k$, we designed a sequential deep learning model that is sensitive to the order of the input.
We will refer to such a model as the \emph{oracle}.

As a standard in sequential deep learning, each operation of a sequence $\pi_k = (s_1^i, s_2^i, \dots, s_{n_i}^i)$ is described by a feature vector in $\mathbb{R}^{g}$.
This means that the representation $X_k$ of a sequence $\pi_k$ is in turn a sequence of feature vectors, or alternatively, a tensor $X_k \in \mathbb{R}^{n_i\times g}$, where the $h \in \{1, \dots, n_i\}$ element describes the operation $s^i_h$.
More information about the features describing an operation is given in Section~\ref{ssec:dataset}.

Our oracle is composed of two blocks: the \emph{first block} takes in the representation of a sequence $X_k$ and creates a \emph{sequence embedding}; the \emph{second block} uses this embedding to output the probability $y_k$ of the sequence.
The entire architecture is depicted in Figure~\ref{fig:architecture}.
\Figure[t!](topskip=0pt, botskip=0pt, midskip=0pt)[width=0.75\textwidth]{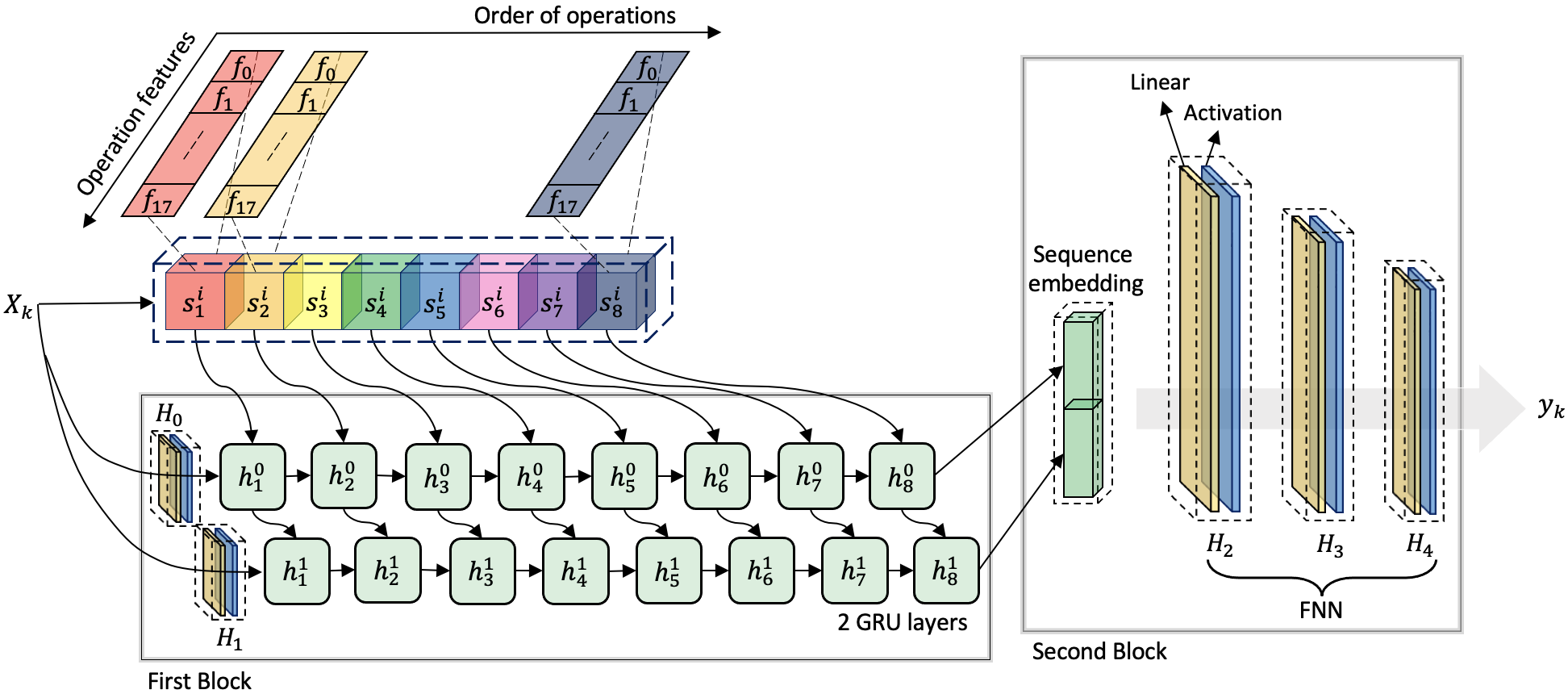}
{The architecture of the oracle. In the left, the 2-dimensional representation $X_k$ of a sequence is transformed into a sequence embedding through the first block. In the right, the sequence embedding is fed into the second block to compute the quality $y_k$.
\label{fig:architecture}}

The first block is realized with two layers of a Gated Recurrent Unit (GRU) \citep{GRU}.
A GRU is a type of Recurrent Neural Network \citep{DeepLearning} that uses a ``memory structure'' to let information from prior inputs influence the current output.
This ``memory structure'' needs to be initialized to some initial state, and is updated at each time step by using the input and current state through a gating mechanism.

Our oracle warms start the initial states with $X_k$, but without considering the order.
Specifically, the initial state of each GRU layer is created by first projecting the feature vectors describing operations in a latent space $\mathbb{R}^{d}$ with a hidden layer ($H_0 \in \mathbb{R}^{g\times d}$ and $H_1 \in \mathbb{R}^{g\times d}$ in Figure~\ref{fig:architecture}), and then by taking the mean along each of the $d$ dimensions.
This allows modeling the concept of a JSP machine directly in the architecture.

After this initialization, starts the creation of the sequence embedding by considering the order of the sequence.
As depicted in Figure~\ref{fig:architecture}, the first GRU layer receives in input at each step $t=(1, \dots, n_i)$ the feature vector of the operation $s_t^i$ and produces in output the state $h_t^0$.
Whereas the second GRU layer receives in input at the step $t$ the state $h_t^0$ and produces in output $h_t^1$.
The final sequence embedding is the concatenation of the last states, $h_{n_i}^0$ and $h_{n_i}^1$, and is therefore a vector in $\mathbb{R}^{2d}$.

The second block is realized with a Feedforward Neural Network (FNN) \citep{DeepLearning} composed by 3 hidden layers of decreasing size.
This block takes in input the sequence embedding and produces in output the probability $y_k$.

\subsection{Tabu Search}\label{ssec:TS}

Since Tabu Search empirically demonstrated to be the best metaheuristic for solving the JSP~\citep{LSA}, we evaluate the advantages of our novel learning task in this algorithm.
To this end, we design two versions of the TS: sTS is a simple TS inspired by the works reviewed in Section~\ref{ssec:LSA}, while oTS is identical to sTS but uses the oracle. 
We borrow part of the structure of sTS and oTS from the TS proposed in \citep{TSAB}.
Since our algorithms are almost identical, they differ only in the procedure to select the next solution, we first describe the structure of sTS and afterwards the modification to the searching procedure. 

The most important blocks of sTS are: 
\begin{enumerate*}[label=(\roman*)]
    \item the generator of the initial solution,
    \item the neighborhood structure,
    \item the tabu list for avoiding revisiting recent solutions,
    \item the neighborhood searching procedure to select the next solution, and
    \item the restart list used to intensify promising regions of the solution space.
\end{enumerate*}

sTS begins by generating a random solution that constitutes both the starting point of the exploration and the initial best solution.
This solution is generated with a random PDR that gives priority to jobs by sampling from a uniform distribution.
We decided to use a random starting point to test the capability of our algorithms to converge to global optima in different runs of the same instance.
This allows a better comparison between the algorithms.

After this initialization, sTS enters the cyclic phase where the following steps are repeated:
\begin{enumerate}[leftmargin=1.5cm,label=\textit{Step} \arabic*:]
    \item \emph{Create the neighborhood} of the current solution.\label{i:N1}
    \item Select the new current solution through the \emph{neighborhood searching procedure}.\label{i:NSP}
    \item \emph{Update the best solution} if the new solution improves the best one.\label{i:update}
    \item Save a restart point in the restart list if the \emph{region is promising}.\label{i:restart}
    \item Go to \ref{i:N1} if the \emph{iteration condition} is met.\label{i:stop1}
    \item Restart from the latest promising region and go to \ref{i:N1} if the \emph{restart condition} is met.\label{i:stop2}
\end{enumerate}

At each iteration, the algorithm selects from the N1 neighborhood~\citep{SA} the solution with the minimum makespan that is not forbidden by the tabu list (\ref{i:NSP}).
Once sTS finds a solution that improves the best one (\ref{i:update}), it records this point in the restart list (\ref{i:restart}).
Based on \citep{TSAB,iTSAB}, a \emph{promising region} of the JSP solution space is a point in which there is an update of the best solution, and such regions must be intensified by trying to explore the entire neighborhood.

This cyclic exploration is repeated until a maximum number of non-improving iterations is reached (\emph{iteration condition} of \ref{i:stop1}), where a non-improving iteration is an iteration that does not improve the best solution. 
If the iteration condition is not met, the algorithm tries to resume the exploration from the last promising region inserted in the restart list.
The \emph{restart condition} of \ref{i:stop2} simply checks that the restart list is not empty. If this condition is not met, the algorithm stops.
The pseudo-code of the neighborhood searching procedure, the tabu list, and the restart list can be found in~\citep{TSAB} 

\smallskip
oTS is identical to sTS, but it uses the oracle to further reduce the N1 neighborhood by excluding solutions that lower the quality of machine permutations.
This aligns with the discussion of Section~\ref{ssec:task}.
Our oracle predicts the likelihood that a sequence (a permutation on some machine $i \in M$) has of belonging to an optimal solution.
In N1, a neighbor solution $\Pi_n$ differs from the current solution $\Pi_c$ in only one permutation on a machine.
Therefore, we use the oracle to remove all the neighbor solutions that introduce a sequence with a lower likelihood of belonging to an optimal solution.
More in detail, if the permutation of $\Pi_n$ on machine $i$ has a higher likelihood of belonging to an optimal solution than $\Pi_c$, we accept this solution in the neighborhood, in the opposite case, we remove $\Pi_n$ from the neighborhood.
The searching procedure for selecting a new solution from this reduced neighborhood remains the same of sTS, that in turn is the same of \citep{TSAB}.
There might be situations in which all the neighbor solutions are removed, in these cases, we undo the reduction and use the normal N1 neighborhood.
Finally, this reduction is applied only for the first quarter of the maximum number of non-improving iterations (\textit{Step} 1-5), and in the same way after every restart.

%% file: sections/4_results.tex
\section{Experimental Results}\label{sec:results}

In this section, we present the dataset used to train and test the neural network oracle of Section~\ref{ssec:oracle}, the results of the oracle on the test set, and the results of our algorithms on 200 JSP instances.

\subsection{The Dataset}\label{ssec:dataset}

We created a dataset of sequences from a set of 200 JSP instances with 8 jobs ($n_i=8, \forall i \in M$) and 8 machines ($m_j=8, \forall j \in J$).
The set of instances has been generated following the guidelines of \citep{benchmarks}.

Then, for each instance, we generated 136 sequences for each machine, and we computed the quality of these sequences with the methodology introduced in Section~\ref{ssec:methodology}.
This results for a single instance $q \in \{1, \dots, 200\}$ in a total of 1088 observations of the form $(X_k^q, y_k^q)$, where $X_k^q \in \mathbb{R}^{n_i\times g}$ is the representation of a machine sequence $\pi_k$, and $y_k^q$ is its quality. 
To ease the notation, in the remainder of this work we omit the index of the instance $q$; nonetheless, remember that each observation of our dataset refers to one and only one instance.

The 136 sequences for each machine have been generated as follows:
\begin{itemize}
    \item 128 \emph{random sequences} by trying to place each operation in all positions of a machine (the pseudocode for generating such sequences is given in Appendix \ref{app:sequences}.).
    \item 1 \emph{optimal sequence} taken from the optimal solution of the instance.
    \item 7 \emph{suboptimal sequences} obtained from the optimal sequence by swapping consecutive operations (we did not swap the first and last operations).
\end{itemize}
The rationale behind these different sequences is that we tried to uniformly sample the characteristics of a machine in an instance. 
The 128 random sequences should reflect the ``unbiased'' impact of the machine on the instance, the optimal sequence is introduced to model the optimality for a machine, and the suboptimal sequences are used to model the neighborhood of an optimal sequence, and hopefully the \emph{Big Valley} phenomenon \citep{iTSAB}.

Regarding the representation $X_k$ of a sequence $\pi_k$, we defined a set of 18 features to describe operations.
Our set of features has been constructed by selecting some of the best features from \citep{Features} and from the graph theory.
The features selected from \citep{Features} describe characteristics about single operations and jobs, some examples are: the processing time of operations and the mean processing time of jobs.
The graph theory features are extracted from the disjunctive graph and they express relationships among operations, some examples are: the eigenvector centrality and the closeness centrality.
These features depend only on information about the instance, therefore, in our experiments, we computed the feature vector for each operation once and we dynamically concatenated the feature vectors in the order given by $\pi_k$ to form $X_k$ ($X_k \in \mathbb{R}^{8 \times 18}$ in this work).
We report in Appendix \ref{app:features} the complete set of features.

This dataset has been used to train and validate the neural network introduced in Section~\ref{ssec:oracle}.

\subsection{Learning Model performance}\label{ssec:NNresults}

\begin{figure}
\includegraphics[width=\columnwidth]{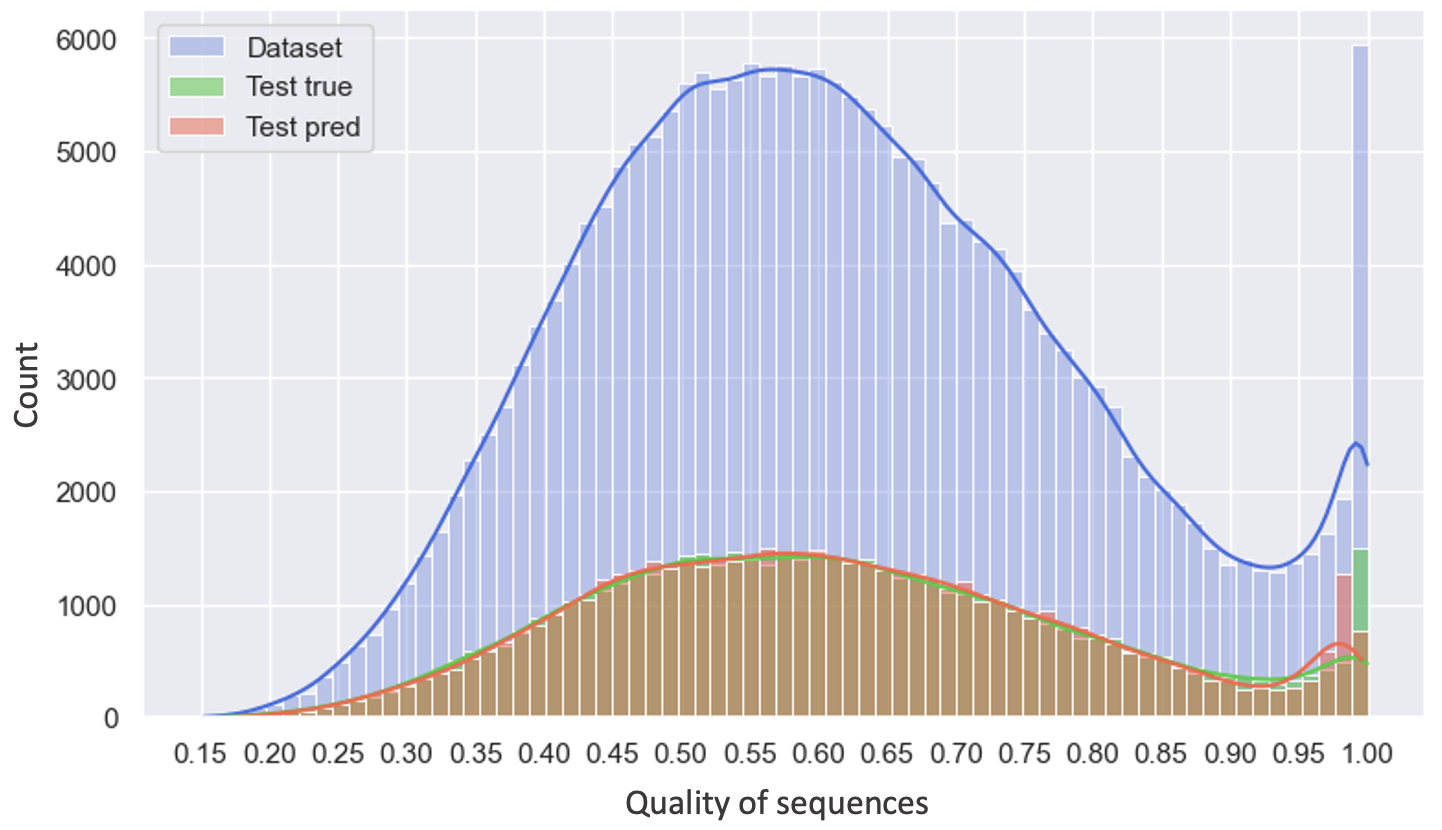}
\caption{The discretized distributions of the dataset (blue), test set (green), and oracle predictions (red).
\label{fig:bin_dist}}
\end{figure}

We evaluate the performance of the oracle on two different aspects:
\begin{enumerate*}[label=(\roman*)]
    \item we quantify the error in the predictions by measuring how much they differ from labels,
    \item we quantify the performance of the oracle in a binary classification problem.
\end{enumerate*}
The results of this section refer to a test set composed of 54400 sequences (25\% of the dataset) randomly selected by ensuring that the test distribution is similar to the one of the entire dataset, see Figure~\ref{fig:bin_dist}.

Due to the nature of our labels $y_k \in [0, 1]$, we trained our oracle to approximate the distribution of the training set by using the Kullback–Leibler Divergence as the loss function.
Using this loss allows to train the model without transforming the problem into a binary classification, and this brings several advantages: 
\begin{enumerate*}[label=(\roman*)]
    \item our labels have a larger semantic compared to binary ones, giving more freedom in the application of the oracle;
    \item it is not clear which threshold should be set on the continuous labels to transform them into binary ones;
    \item casting the problem as a binary classification brings imbalance issues \citep{imbalance}.
\end{enumerate*}
The whole set of hyperparameters and additional training details are given in Appendix~\ref{app:train}.

\smallskip
To quantify the errors of the oracle, we compare its predictions against the labels of the test set by defining the Within Tolerance Accuracy (WTA) in Equation \ref{eq:wta}:\par
{\small
\begin{equation}\label{eq:wta}
    \text{WTA}(tol) = \frac{1}{t}\sum_{k=0}^{t} \mathbb{I}(|y_k - \hat{y_k}| < tol)
\end{equation}
}%
where $\hat{y_k}$ is the predicted quality of a sequence $\pi_k$, $y_k$ is the true quality, $tol$ is the error tolerance, $\mathbb{I}()$ is the indicator function (it returns 1 when the difference is within the tolerance), and $t$ is the dimension of the test set.
\begin{table}
\caption{The errors and WTA for different quality intervals.}
\label{tab:bin_perf}
\begin{center}
\begin{tabular}{K{.12}K{.1}K{.07}K{.07}K{.14}K{.14}}
\toprule
    \multirow{2}{*}{Quality} &
    \multirow{2}{*}{Num} &
    \multicolumn{2}{c}{Abs error} & 
    WTA(.05) & WTA(.07) \\
    \cmidrule(lr){3-4}
    & & avg & max & (\%) & (\%) \\
\midrule
    $[0.0, 0.3)$ & 1117 & 0.012 & 0.078 & 98.0 & 99.7 \\
    $[0.3, 0.4)$ & 4756 & 0.015 & 0.111 & 95.8 & 99.3 \\
    $[0.4, 0.5)$ & 9710 & 0.016 & 0.158 & 94.2 & 98.6 \\
    $[0.5, 0.6)$ & 11775 & 0.018 & 0.150 & 91.7 & 97.8 \\
    $[0.6, 0.7)$ & 10808 & 0.020 & 0.171 & 89.2 & 97.1 \\
    $[0.7, 0.8)$ & 7702 & 0.020 & 0.158 & 87.3 & 96.7 \\
    $[0.8, 0.9)$ & 4343 & 0.021 & 0.152 & 85.9 & 96.2 \\
    $[0.9, 1.0]$ & 4189 & 0.018 & 0.123 & 91.3 & 97.9 \\
\midrule
    \multicolumn{4}{c}{Test set WTA} & 91.0 & 97.7 \\
\bottomrule
\end{tabular}
\end{center}
\end{table}

As it is clear from the distributions in Figure~\ref{fig:bin_dist}, the predictions of the oracle approximate well the distribution of the test set, with some mistakes in the region around the quality 0.90.
This is even more clear from Table~\ref{tab:bin_perf} which quantifies the errors between true and predicted values in different portions of the test distribution.
This table divides the sequences in intervals based on the true quality.
The first column points out the quality intervals, the second column gives the number of occurrences in each interval, the third and fourth columns give some statistics about the absolute error between labels and predictions, and the last two columns give the WTA for different tolerances.
The last row gives the WTA for the entire test set.

Note how the WTA is almost perfect for a tolerance of 0.07, and still very good for a tighter tolerance of 0.05. 
As noted above, we can appreciate an increment in the errors in the interval $[0.7, 0.9)$.
We believe that this increment is jointly caused by the lower number of training sequences in this interval and by the fact that these sequences are more difficult to discriminate from optimal ones because they are mostly suboptimal, i.e., they only differ from optima only in one consecutive pair of operations.

\smallskip
To better understand the quality of the oracle, we also report in Table \ref{tab:binary_perf} its performance in a binary classification task.
In this evaluation, the true quality $y_k$ are transformed into binary labels by setting a threshold and marking with a 1 (positive) all the sequences having a quality higher than the threshold, and with a 0 (negative) all the remaining sequences.
The class predicted by the oracle is given by the \textit{argmax} function.
We report the results for 5 different thresholds, each producing a binary test set with a different imbalance ratio.
Despite these different imbalance ratios, the performance of the oracle on standard imbalanced metrics \citep{imbalance} remains good in all cases.
This is possible because we trained the model to match the quality of the training sequences.

With these evaluations, we want to stress how the oracle can be effectively used either for predicting or classifying sequences, allowing great flexibility in its usage within our TS and potentially in other approximation methods. 
In addition, the results of this section suggest that with our dataset, and hence with the methodology of Section~\ref{ssec:methodology}, it is possible to learn which sequences are likely to be in an optimal solution.

\begin{table*}
\begin{center}
\caption{The results on 5 binary classification problems obtained by setting 5 different thresholds on the labels of the test set.}
\label{tab:binary_perf}
\begin{tabular}{C{.07}C{.07}C{.07}C{.08}C{.07}C{.1}C{.07}C{.06}}
\toprule
Threshold & Num positive & Num negative & Imbalance ratio & Accuracy (\%) & Balanced accuracy (\%) & Precision (\%) & Recall (\%) \\
\midrule
0.5 & 38817 & 15583 & 0.4 & 95.5 & 94.3 & 96.6 & 97.0 \\
0.6 & 27042 & 27358 & 1.01 & 94.7 & 94.7 & 94.6 & 94.8 \\
0.7 & 16234 & 38166 & 2.35 & 95.4 & 94.5 & 92.3 & 92.2 \\
0.8 & 8532  & 45868 & 5.38 & 97.1 & 94.0 & 91.9 & 89.4 \\
0.9 & 4189  & 50211 & 11.99 & 98.6 & 94.4 & 92.2 & 89.4 \\
\bottomrule
\end{tabular}
\end{center}
\end{table*}
\smallskip

\subsection{Tabu Search performance}

We analyze the impact of the proposed learning task by comparing the results of the TS described in Section~\ref{ssec:TS} with (oTS) and without (sTS) the oracle.
This comparison is done on the 200 instances used to create our dataset, where for each instance we repeated the execution of the algorithms 5 times, from the same 5 initial solutions (this is done by seeding the random PDR with 5 different seeds).
The results of the algorithms are compared in terms of the number of optimal solutions, the average optimality gap of suboptimal solutions (gap {\footnotesize $= (C_{max}/C_{max}^{opt}) - 1$} ), and the average execution time.
Note that comparing the results of the algorithms on the same instances of our dataset is fair since the sequences visited by sTS and oTS are independent from those used to train the oracle.

Both the algorithms have been written in C++, compiled with g++ 9.3.0, and executed on an Ubuntu machine equipped with an Intel Core i9-11900K and a NVIDIA GeForce RTX 3090.
Our oracle has been ported from Python by using the tracing functionality of PyTorch \citep{PyTorch}, and it has been integrated in the oTS with LibTorch on the GPU.

\begin{table*}
\caption{The results of the two TS algorithms for different configurations of parameters.}
\label{tab:otsab_perf}
\begin{center}
\begin{tabular}{C{.04}|C{.03}C{.05}|C{.03}C{.03}|C{.03}C{.03}|C{.02}C{.055}|C{.02}C{.055}|C{.03}C{.04}}
\toprule
    \multicolumn{1}{c}{} &
    \multicolumn{2}{c}{Parameters} &
    \multicolumn{2}{c}{Num opt} &
    \multicolumn{2}{c}{Avg opt gap} & 
    \multicolumn{2}{c}{Worse} & 
    \multicolumn{2}{c}{Better} &
    \multicolumn{2}{c}{Avg time} \\
    \cmidrule(lr){2-3} \cmidrule(lr){4-5} \cmidrule(lr){6-7} \cmidrule(lr){8-9} \cmidrule(lr){10-11} \cmidrule(lr){12-13}
    ID & Max iter & Restarts & sTS & oTS & sTS (\%) & oTS (\%) & Num & Avg diff (\%) & Num & Avg diff (\%) & sTS (ms) & oTS (ms)\\ 
\midrule
     0 & 500 & \multirow{5}{*}{0} & 335 & 465 & 2.24 & 2.12 & 277 & 1.56 & 408 & 1.93 & 29 & 142 \\
     1 & 1000 &  & 449 & 620 & 1.81 & 1.77 & 192 & 1.32 & 372 & 1.56 & 41 & 210\\
     2 & 1500 &  & 511 & 673 & 1.66 & 1.41 & 167 & 1.06 & 341 & 1.55 & 56 & 318 \\
     3 & 2000 &  & 559 & 741 & 1.55 & 1.29 & 135 & 1.11 & 320 & 1.49 & 72 & 409 \\
     4 & 2500 &  & 593 & 763 & 1.51 & 1.34 & 116 & 1.10 & 293 & 1.46 & 86 & 473 \\
\midrule
    5 & \multirow{3}{*}{700} 
     & 0 & 382 & 548 & 1.99 & 1.75 & 223 & 1.22 & 406 & 1.75 & 33 & 153 \\
    6 &  & 1 & 741 & 849 & 1.02 & 0.92 & 80 & 0.93 & 195 & 1.02 & 138 & 526 \\
    7 &  & 2 & 807 & 892 & 0.89 & 0.92 & 65 & 0.97 & 150 & 0.89 & 199 & 690 \\
\midrule
    8 & \multirow{3}{*}{800} 
     & 0 & 409 & 572 & 1.89 & 1.71 & 211 & 1.28 & 395 & 1.65 & 35 & 165 \\
    9 & & 1 & 754 & 867 & 1.01 & 0.84 & 71 & 0.79 & 185 & 1.04 & 181 & 592 \\
    10 &  & 2 & 818 & 905 & 0.89 & 0.85 & 58 & 0.80 & 140 & 0.91 & 245 & 795 \\
\midrule
    11 & \multirow{3}{*}{900} 
     & 0 & 425 & 594 & 1.84 & 1.69 & 218 & 1.24 & 377 & 1.70 & 38 & 191 \\
    12 &  & 1 & 766 & 879 & 1.00 & 0.97 & 66 & 0.96 & 173 & 1.04 & 204 & 714 \\
    13 &  & 2 & 833 & 918 & 0.88 & 0.82 & 48 & 0.93 & 128 & 0.97 & 263 & 961 \\
\bottomrule
\end{tabular}
\end{center}
\end{table*}

In Table~\ref{tab:otsab_perf}, we report the results of the algorithms for different configurations of parameters.
In these configurations, we omit the length of the tabu list that is always set to 10.
The first column of the table assigns an identifier to every configuration. 
The second and third column specifies respectively the maximum number of non-improving iterations and the length of the restart list.
The ``\textit{Num opt}'' and the ``\textit{Avg opt gap}'' columns compare the number of optimal solutions and the average optimality gap of each algorithm.
Whereas the ``\textit{Worse}'' (``\textit{Better}'') columns compare respectively the number of solutions and the average scaled difference (diff {\footnotesize $= (C_{max}^{oTS} - C_{max}^{sTS}) / C_{max}^{opt}$}) in which oTS worsens (improves) with respects to sTS. 
The last two columns give the average execution time of each algorithm.

First, we want to underline that oTS finds a higher number of optimal solutions than sTS regardless of the parameter configurations.
This is important for empirically confirming that the proposed learning task, our methodology, and the learning model indeed enhance the performance of the TS.

This increment in performance is also supported by the lower average optimality gaps obtained by oTS in suboptimal solutions (``\textit{Avg opt gap}'' columns).
For all the tested configurations, we only see one case, row with ID 7, in which oTS does slightly worse than sTS in terms of optimality gap.
However, note how in this case the overall performance of both the algorithms is almost perfect, and how oTS is still able to find a larger number of optimal solutions.

Regarding the ``\textit{Worse}'' and the ``\textit{Better}'' columns, we just highlight how the number of solutions in which oTS does better than sTS is almost twice the number of solutions in which it does worse.

Finally, as it is obvious from the average times, using a deep learning model will likely increase the running time.
This trend has already been observed for instance in \citep{DispatchDRL}, where the execution of their DRL proposal takes 2x up to 5x the time of traditional PDRs.
A similar increment is also observed in \citep{AC-DRL}.
In line with these works, we observe a comparable increment between sTS and oTS.
However, our algorithms have been written by keeping the implementation as simple as possible.
Therefore, there is space for engineering the code and producing better average execution times, especially in the case of the oTS.
For instance, it is possible to reduce the calls to the oracle by batching or keeping a memory of past predictions, and it is possible to reduce the execution time of the oracle by using faster architectures like Transformers \citep{Transformer} and Convolutional Neural Network \citep{DeepLearning}.\par

Concluding, this comparative analysis shows that it is possible to find better solutions by using the quality predictions in a TS as described in Section~\ref{ssec:TS}. 
This empirically highlights how the proposed learning task seems to be valuable in the context of the JSP.

\subsection{Comparison with Reinforcement Learning}

In this section, we compare the performance of oTS with the proposals relying on DRL.
The objective is to justify our efforts by demonstrating the superiority of metaheuristics enhanced with machine learning and the importance of further investigating these hybrid approaches.

For this comparison, we selected instances from the works discussed in Section~\ref{ssec:PDR}.
Specifically, we selected the instances Orb01-09~\citep{Orb} and the instances Ta01-10~\citep{benchmarks}.
We report in Table~\ref{tab:drl_comp} the instance name, the optimal makespan, and the results in terms of makespan and optimality gap (in round brackets) for the Shortest Processing Time (SPT), the proposal of \citep{AC-DRL}, the proposal of \citep{DispatchDRL}, and oTS.
Based on Table~\ref{tab:otsab_perf}, we decided to use 2 parameter configurations for oTS: 2 restarts and 700 iterations for oTS-1, and 2 restarts and 800 iterations for oTS-2.

\begin{table*}
\begin{center}
\caption{The comparison between oTS and the DRL approaches on benchmark instances.}
\label{tab:drl_comp}
\begin{tabular}{C{.005}C{.04}|R{.03}R{.09}|R{.09}R{.09}|R{.08}R{.08}}
\toprule
    \multicolumn{2}{c}{Instance} & 
    \centering OPT & 
    \centering SPT & 
    \centering \citep{AC-DRL} & 
    \centering \citep{DispatchDRL} &
    \centering oTS-1 & 
    oTS-2 \;\;\;\; \\
\midrule
     \multirow{9}{*}{\rotatebox{90}{$10 \times 10 $}} 
     & Orb01 & 1059 & 1478\;(39.6\%) & 1211\;(14.4\%) & - \;\;\;\;\;\;\; & 1106\;(4.4\%) & 1106\;(4.4\%) \\
     & Orb02 &  888 & 1175\;(32.3\%) & 1002\;(12.8\%) & - \;\;\;\;\;\;\; &  902\;(1.6\%) &  902\;(1.6\%) \\
     & Orb03 & 1005 & 1179\;(17.3\%) & 1150\;(14.4\%) & - \;\;\;\;\;\;\; & 1048\;(4.3\%) & 1044\;(3.9\%) \\
     & Orb04 & 1005 & 1236\;(23.0\%) & 1132\;(12.6\%) & - \;\;\;\;\;\;\; & 1032\;(2.7\%) & 1032\;(2.7\%) \\
     & Orb05 &  887 & 1152\;(29.9\%) & 1045\;(17.8\%) & - \;\;\;\;\;\;\; &  902\;(1.7\%) &  896\;(1.0\%) \\
     & Orb06 & 1010 & 1190\;(17.8\%) & 1106\;(9.5\%) & - \;\;\;\;\;\;\; & 1028\;(1.8\%) & 1028\;(1.8\%) \\
     & Orb07 &  397 &  504\;(27.0\%) &  460\;(15.9\%) & - \;\;\;\;\;\;\; & 397\;(0.0\%) &  397\;(0.0\%) \\
     & Orb08 &  899 & 1170\;(30.1\%) & 1022\;(13.7\%) & - \;\;\;\;\;\;\; & 911\;(1.3\%) &  911\;(1.3\%) \\
     & Orb09 &  934 & 1262\;(35.1\%) & 1082\;(15.8\%) & - \;\;\;\;\;\;\; &  961\;(2.9\%) &  955\;(2.2\%) \\
 \midrule
     \multirow{10}{*}{\rotatebox{90}{$15 \times 15 $}} 
     & Ta01 & 1231 & 1872\;(52.1\%)  & - \;\;\;\;\;\;\; & 1443\;(17.2\%) & 1281\;(4.1\%) & 1281\;(4.1\%) \\
     & Ta02 & 1244 & 1709\;(37.4\%)  & - \;\;\;\;\;\;\; & 1544\;(24.1\%) & 1283\;(3.1\%) & 1283\;(3.1\%)\\
     & Ta03 & 1218 & 2009\;(64.9\%)  & - \;\;\;\;\;\;\; & 1440\;(18.2\%) & 1292\;(6.1\%) & 1292\;(6.1\%)\\
     & Ta04 & 1175 & 1825\;(55.3\%)  & - \;\;\;\;\;\;\; & 1637\;(39.3\%) & 1248\;(6.2\%) & 1248\;(6.2\%)\\
     & Ta05 & 1224 & 2044\;(67.0\%)  & - \;\;\;\;\;\;\; & 1619\;(32.3\%) & 1280\;(4.6\%) & 1280\;(4.6\%)\\
     & Ta06 & 1238 & 1771\;(43.1\%)  & - \;\;\;\;\;\;\; & 1601\;(29.3\%) & 1272\;(2.7\%) & 1260\;(1.8\%)\\
     & Ta07 & 1227 & 2016\;(64.3\%)  & - \;\;\;\;\;\;\; & 1568\;(27.8\%) & 1250\;(1.9\%) & 1247\;(1.6\%)\\
     & Ta08 & 1217 & 1654\;(35.9\%)  & - \;\;\;\;\;\;\; & 1468\;(20.6\%) & 1240\;(1.9\%) & 1240\;(1.9\%)\\
     & Ta09 & 1274 & 1962\;(54.0\%)  & - \;\;\;\;\;\;\; & 1627\;(27.7\%) & 1307\;(2.6\%) & 1307\;(2.6\%)\\
     & Ta10 & 1241 & 2164\;(74.4\%)  & - \;\;\;\;\;\;\; & 1527\;(23.0\%) & 1290\;(3.9\%) & 1290\;(3.9\%) \\
\bottomrule
\end{tabular}
\end{center}
\end{table*}

From Table~\ref{tab:drl_comp}, it is immediately clear that oTS outperforms the DRL proposals.
This is also true if we qualitatively compare the results of oTS with those reported in \citep{ScheduleDRL}.
By looking at the average percentage gap reported for Orb01-10 and Ta01-80, we can see that the gap of this other DRL proposal is around 20\%, ten times the gap obtained by oTS.

This comparison demonstrates that metaheuristics enhanced with machine learning guarantee to find better solutions.
We believe that further research in hybrid approaches as our may give life to simpler and better metaheuristics capable of producing near-optimal solutions in a shorter amount of time.

%% file: sections/5_close.tex
\section{Conclusions}\label{sec:close}

In this work, we proposed a novel supervised learning task for the JSP that aims at predicting the quality of machine permutations.
We designed an original methodology to estimate this quality by means of a MILP solver.
Then, we constructed a dataset with this methodology, and we demonstrated that is possible to learn a flexible and accurate sequential deep learning model to predict the quality of machine permutations.
Finally, we justified both theoretically and empirically the benefits of using the proposed learning task in the context of metaheuristics for the JSP.

Our aim was to propose a simple and reasonable methodology that allows evaluating the benefits of applying supervised learning to the JSP.
Although DRL seems a more natural and established ML paradigm for this problem, our analysis suggests that also 
supervised learning is a valuable and viable paradigm, especially if used in tandem with existing approximation methods.

In future works, we will address the main limitation of our hybrid metaheuristic: the increase in the execution time of the algorithm.
We will investigate whether our learning task could benefit other methods for solving the JSP, like machine-based decomposition and ruin-and-recreate algorithms, and whether it is possible to develop new ad-hoc methods.  
In addition, we believe there is a need to extensively compare the benefits and drawbacks of the new ML-based proposals with a wide spectrum of well-established algorithms.

%% file: sections/AppA.tex
\section{sequence generator}\label{app:sequences}

The sequence generator procedure takes in a sequence of operations on some machine $i \in M$ and generates $s$ random sequences.
This procedure is applied on each machine of an instance, and it tries to place each operation in all the $n_i$ positions of a permutation.
Note that this procedure may generate repeated sequences.
Such repeated sequences must be removed, and the procedure must be called again to ensure that $s$ different sequences are generated. 
The symbol $\oplus$ indicates that an item is appended to a partial sequence.

\smallskip
\begin{algorithmic}
\Function{SequenceGenerator}{$( s_1^i, \dots, s_{n_i}^i ), s$}
\State $seq \gets$ Generate $s$ empty sequences
\State $w \gets ( s_1^i, \dots, s_{n_i}^i )$
\ForAll{$pos \in \{ 1, \dots, n_i \}$}
    \State $idx = 0$
    \ForAll{$k \in \{ 0, \dots, s - 1 \}$}
        \While{$w_{idx}$ in $seq_k$}
            \State $idx = (idx + 1) \text{mod}\, |w|$
        \EndWhile
        \State $seq_k \gets seq_k \oplus w_{idx}$
        \State $idx = (idx + 1) \text{mod}\, |w|$
    \EndFor
    \State $w \gets w \oplus ( s_1^i, \dots, s_n^i )$
    \Comment{Increase $w$'s period.}
    \State $w \gets \text{shuffle}(w)$
\EndFor
\State \Return seq
\EndFunction
\end{algorithmic}

%% file: sections/AppB.tex
\section{hyperparameters and training details}\label{app:train}

Table~\ref{tab:hyperparams} reports the hyperparameter of the sequential deep learning model of Section~\ref{ssec:oracle}. 
This model has been trained with the adam optimizer~\citep{Adam}, with a batch size of 128, and for a total of 100 epochs divided as follows:
\begin{enumerate}
    \item 40 epochs with learning rate 0.005.
    \item 30 epochs with learning rate 0.002.
    \item 20 epochs with learning rate 0.001.
    \item 10 epochs with learning rate 0.0005.
\end{enumerate}

\begin{table}[!htbp]
\centering
\caption{Hyperparameters.}
\arrayrulecolor{black}
    \begin{tabular}{L{.12}L{.3}L{.08}}
    \toprule
    Block & hyperparameter & Value \\
    \midrule
    \multirow{6}{*}{GRU 0} & hidden size & 32 \\
     & bidirectional & False \\
     & dropout & 0.3 \\
     & $H_0$ size & 32 \\
     & $H_0$ activation & $\tanh$ \\
     & $H_0$ dropout & 0.3 \\
     \midrule
     \multirow{5}{*}{GRU 1} & hidden size & 32 \\
     & bidirectional & False \\
     & $H_1$ size & 32 \\
     & $H_1$ activation & $\tanh$ \\
     & $H_1$ dropout & 0.3 \\
    \midrule
    \multirow{5}{*}{FNN} & $H_2$ size & 32 \\
     & $H_2$ activation & $\tanh$ \\
     & $H_3$ size & 16 \\
     & $H_3$ activation & $\tanh$ \\
     & $H_4$ size & 2 \\
    \bottomrule
    \end{tabular}
\label{tab:hyperparams}
\end{table}

%% file: sections/AppC.tex
\section{features}\label{app:features}

Table~\ref{tab:features} reports the set of 18 features used to describe operations in this work.
The first column gives to each feature a unique identifier (in accordance with Figure~\ref{fig:architecture}), the second column points out the feature name, the third the equation, and the last column a brief description about the feature.
The graph theory features (from row $f_8$ to row $f_{17}$) have been computed with the NetworkX package.

\begin{table*}[htbp]
\begin{center}
\caption{The set of features describing an operation and its relations with others in the JSP instance.}
\label{tab:features}
\resizebox{0.9\textwidth}{!}{
\begin{tabular}{C{.05}C{.10}C{.15}L{.9}}
\toprule
Feature & Name & Equation & Description \\
\midrule
$f_0$ & Processing time & $\frac{p_j^i}{\max_{ik} p_k^i}$ & The processing time of operation $o^i_j$ normalized by the maximum processing time in the instance. \\ 
\midrule
$f_1$ & Job completion & $\frac{\sum_{k=1}^{i} p_j^k}{\sum_{k=1}^{m_j} p_j^k}$ & The completion of job $j$ when its operation $o^i_j$ is scheduled. \\
\midrule
$f_2$ & Job mean & $\frac{1}{\text{avg}\,*\,m_j} \sum_{i=1}^{m_j} p_j^i$ & Mean processing time of job $j$ normalized by the mean processing time of the instance ($\text{avg}$ in the equation). \\
\midrule
$f_3$ & Job median & $\frac{\text{median}(p_j^1, \dots, p_j^{m_j})}{\text{avg}}$ & Median processing time of job $j$ scaled by the mean processing time of the instance ($\text{avg}$ in the equation). \\
\midrule
$f_4$ & Job std-mean & $\frac{\text{std}(p_j^1, \dots, p_j^{m_j})\,m_j}{\sum_{i=1}^{m_j} p_j^i}$ &  The standard deviation of the processing time in job $j$ normalized by the mean processing time of the job. \\
\midrule
$f_5$ & Job std-median & $\frac{\text{std}(p_j^1, \dots, p_j^{m_j})}{\text{median}(p_j^1, \dots, p_j^{m_j})}$ &  The standard deviation of the processing time in job $j$ normalized by the median processing time of the job. \\
\midrule
$f_6$ & Job min & $\frac{\min_{i \in \{ 1, \dots, m_j \}} p_j^i}{\max_{ik} p_k^i}$ & The minimum processing time of job $j$ normalized by the maximum processing time in the instance. \\
\midrule
$f_7$ & Job max & $\frac{\max_{i \in \{ 1, \dots, m_j \}} p_j^i}{\max_{ik} p_k^i}$ & The maximum processing time of job $j$ normalized by the maximum processing time in the instance. \\
\midrule
$f_{8}$ & Source shortest distance & $d^*(\text{src}, o^i_j)$ & The shortest weighted distance in the graph from the source (dummy) node to operation $o^i_j$. \\
\midrule
$f_{9}$ & Destination shortest distance & $d^*(o^i_j, \text{dst})$ & The shortest weighted distance in the graph from operation $o^i_j$ to the destination (dummy) node. \\
\midrule
$f_{10}$ & Eigenvector Centrality & $Ax = x\lambda$ & The eigenvector centrality of an operation $o^i_j$ is the element of the eigenvector $x$ associated with the largest eigenvalue $\lambda$ that corresponds to $o^i_j$.
A high eigenvector centrality means that an operation connects to other operations having high centrality.
$A$ is the adjacency matrix of the graph. \\
\midrule
$f_{11}$ & Weighted Eigenvector Centrality & $A^{*}x = x\lambda$ & The same as the eigenvector centrality, but it uses the weighted adjacency matrix $A^{*}$ where arcs take the weight of the source node. \\
\midrule
$f_{12}$ & Closeness Centrality & $\frac{|V| - 1}{ \sum_{k \in \mathcal{N}(o^i_j)} d(k, o^i_j)}$ & The normalized closeness centrality measures the shortest non-weighted distance from the nodes than can reach $o^i_j$, scaled by the number of nodes in the graph. $\mathcal{N}(o^i_j)$ is the set of nodes that can reach $o^i_j$, $d(k, o^i_j)$ is the number of arcs on the shortest path from $k$ to $o^i_j$, and $|V|$ is the number of nodes.\\
\midrule
$f_{13}$ & Weighted Closeness Centrality & $\frac{|V| - 1}{\sum_{k \in \mathcal{N}(o^i_j)} d^*(k, \, o^i_j)}$ & The same as the closeness centrality, but it uses the weighted shortest path $d^*(k, o^i_j)$. \\
\midrule
$f_{14}$ & Betweenness Centrality & $\sum_{v,w \in V} \frac{\Gamma_{v \rightarrow w}(o^i_j)}{\Gamma_{v \rightarrow w}}$ & The betweenness centrality is the fraction of all-pairs shortest paths that pass through operation $o^i_j$. 
This measure indicates which operations are ``bridges'' between others in a graph.
$\Gamma_{v \rightarrow w}$ is the number of non-weighted shortest paths from $v$ to $w$, and $\Gamma_{v \rightarrow w}(o^i_j)$ is the number of such shortest paths through $o^i_j$.\\
\midrule
$f_{15}$ & Weighted Betweenness Centrality & $\sum_{v,w \in V} \frac{\Gamma^*_{v \rightarrow w}(o^i_j)}{\Gamma^*_{v \rightarrow w}}$ & The same as the betweenness centrality, but it uses the weighted shortest path for computing the number of paths $\Gamma^*_{v \rightarrow w}$. \\
\midrule
$f_{16}$ & Page Rank & $A$ & The Page Rank. \\
\midrule
$f_{17}$ & Weighted Page Rank & $A^*$ & The weighted Page Rank. \\
\bottomrule
\end{tabular}
}
\end{center}
\end{table*}

%% file: main.bbl
\begin{thebibliography}{44}
\providecommand{\natexlab}[1]{#1}
\providecommand{\url}[1]{\texttt{#1}}
\expandafter\ifx\csname urlstyle\endcsname\relax
  \providecommand{\doi}[1]{doi: #1}\else
  \providecommand{\doi}{doi: \begingroup \urlstyle{rm}\Url}\fi

\bibitem[Zhang et~al.(2019)Zhang, Ding, Zou, Qin, and Fu]{JSP4.0}
Jian Zhang, Guofu Ding, Yisheng Zou, Shengfeng Qin, and Jianlin Fu.
\newblock Review of job shop scheduling research and its new perspectives under
  industry 4.0.
\newblock \emph{Journal of Intelligent Manufacturing}, 30:\penalty0 1809--1830,
  2019.

\bibitem[Bello et~al.(2017)Bello, Pham, Le, Norouzi, and Bengio]{TSP}
Irwan Bello, Hieu Pham, Quoc~V Le, Mohammad Norouzi, and Samy Bengio.
\newblock Neural combinatorial optimization with reinforcement learning.
\newblock \emph{International Conference on Learning Representations}, 2017.

\bibitem[Khalil et~al.(2017)Khalil, Dai, Zhang, Dilkina, and Song]{GNNDRL}
Elias Khalil, Hanjun Dai, Yuyu Zhang, Bistra Dilkina, and Le~Song.
\newblock Learning combinatorial optimization algorithms over graphs.
\newblock \emph{Advances in neural information processing systems}, 30, 2017.

\bibitem[Nazari et~al.(2018)Nazari, Oroojlooy, Snyder, and Tak{\'a}c]{VRP}
Mohammadreza Nazari, Afshin Oroojlooy, Lawrence Snyder, and Martin Tak{\'a}c.
\newblock Reinforcement learning for solving the vehicle routing problem.
\newblock \emph{Advances in neural information processing systems}, 31, 2018.

\bibitem[Liu et~al.(2020)Liu, Chang, and Tseng]{AC-DRL}
Chien-Liang Liu, Chuan-Chin Chang, and Chun-Jan Tseng.
\newblock Actor-critic deep reinforcement learning for solving job shop
  scheduling problems.
\newblock \emph{IEEE Access}, 8:\penalty0 71752--71762, 2020.

\bibitem[Zhang et~al.(2020)Zhang, Song, Cao, Zhang, Tan, and Chi]{DispatchDRL}
Cong Zhang, Wen Song, Zhiguang Cao, Jie Zhang, Puay~Siew Tan, and Xu~Chi.
\newblock Learning to dispatch for job shop scheduling via deep reinforcement
  learning.
\newblock \emph{Advances in Neural Information Processing Systems},
  33:\penalty0 1621--1632, 2020.

\bibitem[Park et~al.(2021)Park, Chun, Kim, Kim, and Park]{ScheduleDRL}
Junyoung Park, Jaehyeong Chun, Sang Kim, Youngkook Kim, and Jinkyoo Park.
\newblock Learning to schedule job-shop problems: representation and policy
  learning using graph neural network and reinforcement learning.
\newblock \emph{International Journal of Production Research}, 59:\penalty0
  1--18, 01 2021.

\bibitem[Bengio et~al.(2021)Bengio, Lodi, and Prouvost]{Horizon}
Yoshua Bengio, Andrea Lodi, and Antoine Prouvost.
\newblock Machine learning for combinatorial optimization: A methodological
  tour d’horizon.
\newblock \emph{European Journal of Operational Research}, 290\penalty0
  (2):\penalty0 405--421, 2021.
\newblock ISSN 0377-2217.

\bibitem[Mazyavkina et~al.(2021)Mazyavkina, Sviridov, Ivanov, and
  Burnaev]{RLCOSurvey}
Nina Mazyavkina, Sergey Sviridov, Sergei Ivanov, and Evgeny Burnaev.
\newblock Reinforcement learning for combinatorial optimization: A survey.
\newblock \emph{Computers \& Operations Research}, 134:\penalty0 105400, 2021.
\newblock ISSN 0305-0548.

\bibitem[Pinedo(2012)]{Scheduling}
Michael~L Pinedo.
\newblock \emph{Scheduling}, volume~29.
\newblock Springer, 2012.

\bibitem[Ku and Beck(2016)]{MIP}
Wen-Yang Ku and J.~Christopher Beck.
\newblock Mixed integer programming models for job shop scheduling: A
  computational analysis.
\newblock \emph{Computers \& Operations Research}, 73:\penalty0 165--173, 2016.
\newblock ISSN 0305-0548.

\bibitem[Sutton and Barto(2018)]{Sutton}
Richard~S Sutton and Andrew~G Barto.
\newblock \emph{Reinforcement learning: An introduction}.
\newblock MIT press, 2018.

\bibitem[Henderson et~al.(2018)Henderson, Islam, Bachman, Pineau, Precup, and
  Meger]{DRLRepro}
Peter Henderson, Riashat Islam, Philip Bachman, Joelle Pineau, Doina Precup,
  and David Meger.
\newblock Deep reinforcement learning that matters.
\newblock \emph{Proceedings of the AAAI Conference on Artificial Intelligence},
  32\penalty0 (1), Apr. 2018.

\bibitem[Adams et~al.(1988)Adams, Balas, and Zawack]{bottleneck}
Joseph Adams, Egon Balas, and Daniel Zawack.
\newblock The shifting bottleneck procedure for job shop scheduling.
\newblock \emph{Management science}, 34\penalty0 (3):\penalty0 391--401, 1988.

\bibitem[Haupt(1989)]{PDR}
Reinhard Haupt.
\newblock A survey of priority rule-based scheduling.
\newblock \emph{Operations-Research-Spektrum}, 11\penalty0 (1):\penalty0 3--16,
  1989.

\bibitem[Mouelhi-Chibani and Pierreval(2010)]{PDRNN}
Wiem Mouelhi-Chibani and Henri Pierreval.
\newblock Training a neural network to select dispatching rules in real time.
\newblock \emph{Computers \& Industrial Engineering}, 58\penalty0 (2):\penalty0
  249--256, 2010.
\newblock ISSN 0360-8352.
\newblock Scheduling in Healthcare and Industrial Systems.

\bibitem[Ingimundardottir and Runarsson(2018)]{ImitationPDR}
Helga Ingimundardottir and Thomas~Philip Runarsson.
\newblock Discovering dispatching rules from data using imitation learning: A
  case study for the job-shop problem.
\newblock \emph{J. of Scheduling}, 21\penalty0 (4):\penalty0 413–428, aug
  2018.
\newblock ISSN 1094-6136.

\bibitem[Konda and Tsitsiklis(1999)]{ActorCritic}
Vijay Konda and John Tsitsiklis.
\newblock Actor-critic algorithms.
\newblock \emph{Advances in neural information processing systems}, 12, 1999.

\bibitem[Wu et~al.(2021)Wu, Pan, Chen, Long, Zhang, and Yu]{GNN}
Zonghan Wu, Shirui Pan, Fengwen Chen, Guodong Long, Chengqi Zhang, and
  Philip~S. Yu.
\newblock A comprehensive survey on graph neural networks.
\newblock \emph{IEEE Transactions on Neural Networks and Learning Systems},
  32\penalty0 (1):\penalty0 4–24, Jan 2021.
\newblock ISSN 2162-2388.

\bibitem[Talbi(2009)]{Metaheuristic}
El-Ghazali Talbi.
\newblock \emph{Metaheuristics: from design to implementation}, volume~74.
\newblock John Wiley \& Sons, 2009.

\bibitem[Aarts et~al.(1994)Aarts, van Laarhoven, Lenstra, and Ulder]{LSA}
Emile~HL Aarts, Peter~JM van Laarhoven, Jan~Karel Lenstra, and Nico~LJ Ulder.
\newblock A computational study of local search algorithms for job shop
  scheduling.
\newblock \emph{ORSA Journal on Computing}, 6\penalty0 (2):\penalty0 118--125,
  1994.

\bibitem[van Laarhoven et~al.(1992)van Laarhoven, Aarts, and Lenstra]{SA}
Peter J.~M. van Laarhoven, Emile H.~L. Aarts, and Jan~Karel Lenstra.
\newblock Job shop scheduling by simulated annealing.
\newblock \emph{Operations Research}, 40\penalty0 (1):\penalty0 113--125, 1992.
\newblock ISSN 0030364X, 15265463.

\bibitem[Kirkpatrick et~al.(1983)Kirkpatrick, Gelatt, and Vecchi]{firstSA}
S.~Kirkpatrick, C.~D. Gelatt, and M.~P. Vecchi.
\newblock Optimization by simulated annealing.
\newblock \emph{SCIENCE}, 220\penalty0 (4598):\penalty0 671--680, 1983.

\bibitem[Dell'Amico and Trubian(1993)]{TABU}
Mauro Dell'Amico and Marco Trubian.
\newblock Applying tabu search to the job-shop scheduling problem.
\newblock \emph{Ann. Oper. Res.}, 41\penalty0 (1–4):\penalty0 231–252, may
  1993.
\newblock ISSN 0254-5330.

\bibitem[Nowicki and Smutnicki(1996)]{TSAB}
Eugeniusz Nowicki and Czeslaw Smutnicki.
\newblock A fast taboo search algorithm for the job shop problem.
\newblock \emph{Management Science}, 42\penalty0 (6):\penalty0 797--813, 1996.
\newblock ISSN 00251909, 15265501.

\bibitem[Zhang et~al.(2007)Zhang, Li, Guan, and Rao]{N7}
ChaoYong Zhang, PeiGen Li, ZaiLin Guan, and YunQing Rao.
\newblock A tabu search algorithm with a new neighborhood structure for the job
  shop scheduling problem.
\newblock \emph{Computers \& Operations Research}, 34\penalty0 (11):\penalty0
  3229--3242, 2007.
\newblock ISSN 0305-0548.

\bibitem[Glover and Laguna(1998)]{TS}
Fred Glover and Manuel Laguna.
\newblock \emph{Tabu Search}, pages 2093--2229.
\newblock Springer US, Boston, MA, 1998.
\newblock ISBN 978-1-4613-0303-9.

\bibitem[Nowicki and Smutnicki(2005)]{iTSAB}
Eugeniusz Nowicki and Czeslaw Smutnicki.
\newblock An advanced tabu search algorithm for the job shop problem.
\newblock \emph{Journal of Scheduling}, 8:\penalty0 145--159, 2005.

\bibitem[Huang and Liao(2008)]{AATS}
Kuo-Ling Huang and Ching-Jong Liao.
\newblock Ant colony optimization combined with taboo search for the job shop
  scheduling problem.
\newblock \emph{Computers \& Operations Research}, 35\penalty0 (4):\penalty0
  1030--1046, 2008.
\newblock ISSN 0305-0548.

\bibitem[Sha and Hsu(2006)]{PSOTS}
D.Y. Sha and Cheng-Yu Hsu.
\newblock A hybrid particle swarm optimization for job shop scheduling problem.
\newblock \emph{Computers \& Industrial Engineering}, 51\penalty0 (4):\penalty0
  791--808, 2006.
\newblock ISSN 0360-8352.

\bibitem[Cheng et~al.(1999)Cheng, Gen, and Tsujimura]{SurveyGA}
Runwei Cheng, Mitsuo Gen, and Yasuhiro Tsujimura.
\newblock A tutorial survey of job-shop scheduling problems using genetic
  algorithms, part ii: hybrid genetic search strategies.
\newblock \emph{Computers \& Industrial Engineering}, 36\penalty0 (2):\penalty0
  343--364, 1999.

\bibitem[Talbi(2021)]{MLintoHeuristics}
El-Ghazali Talbi.
\newblock Machine learning into metaheuristics: A survey and taxonomy.
\newblock \emph{ACM Computing Surveys (CSUR)}, 54\penalty0 (6):\penalty0 1--32,
  2021.

\bibitem[Chen and Tian(2019)]{LocalRewriting}
Xinyun Chen and Yuandong Tian.
\newblock Learning to perform local rewriting for combinatorial optimization.
\newblock \emph{Advances in Neural Information Processing Systems}, 32, 2019.

\bibitem[Thevenin and Zufferey(2019)]{LearningVNS}
Simon Thevenin and Nicolas Zufferey.
\newblock Learning variable neighborhood search for a scheduling problem with
  time windows and rejections.
\newblock \emph{Discrete Applied Mathematics}, 261:\penalty0 344--353, 2019.
\newblock ISSN 0166-218X.
\newblock GO X Meeting, Rigi Kaltbad (CH), July 10--14, 2016.

\bibitem[Schrimpf et~al.(2000)Schrimpf, Schneider, Stamm-Wilbrandt, and
  Dueck]{RR}
Gerhard Schrimpf, Johannes Schneider, Hermann Stamm-Wilbrandt, and Gunter
  Dueck.
\newblock Record breaking optimization results using the ruin and recreate
  principle.
\newblock \emph{J. Comput. Phys.}, 159\penalty0 (2):\penalty0 139–171, apr
  2000.
\newblock ISSN 0021-9991.

\bibitem[Chung et~al.(2014)Chung, Gulcehre, Cho, and Bengio]{GRU}
Junyoung Chung, Caglar Gulcehre, Kyunghyun Cho, and Yoshua Bengio.
\newblock Empirical evaluation of gated recurrent neural networks on sequence
  modeling.
\newblock In \emph{NIPS 2014 Workshop on Deep Learning, December 2014}, 2014.

\bibitem[Goodfellow et~al.(2016)Goodfellow, Bengio, and
  Courville]{DeepLearning}
Ian Goodfellow, Yoshua Bengio, and Aaron Courville.
\newblock \emph{Deep Learning}.
\newblock MIT Press, 2016.
\newblock \url{http://www.deeplearningbook.org}.

\bibitem[Taillard(1993)]{benchmarks}
E.~Taillard.
\newblock Benchmarks for basic scheduling problems.
\newblock \emph{European Journal of Operational Research}, 64\penalty0
  (2):\penalty0 278--285, 1993.
\newblock ISSN 0377-2217.
\newblock Project Management anf Scheduling.

\bibitem[Mirshekarian and Šormaz(2016)]{Features}
Sadegh Mirshekarian and Dušan~N. Šormaz.
\newblock Correlation of job-shop scheduling problem features with scheduling
  efficiency.
\newblock \emph{Expert Systems with Applications}, 62:\penalty0 131--147, 2016.
\newblock ISSN 0957-4174.

\bibitem[He and Garcia(2009)]{imbalance}
Haibo He and Edwardo~A Garcia.
\newblock Learning from imbalanced data.
\newblock \emph{IEEE Transactions on knowledge and data engineering},
  21\penalty0 (9):\penalty0 1263--1284, 2009.

\bibitem[Paszke et~al.(2019)Paszke, Gross, Massa, Lerer, Bradbury, Chanan,
  Killeen, Lin, Gimelshein, Antiga, Desmaison, Kopf, Yang, DeVito, Raison,
  Tejani, Chilamkurthy, Steiner, Fang, Bai, and Chintala]{PyTorch}
Adam Paszke, Sam Gross, Francisco Massa, Adam Lerer, James Bradbury, Gregory
  Chanan, Trevor Killeen, Zeming Lin, Natalia Gimelshein, Luca Antiga, Alban
  Desmaison, Andreas Kopf, Edward Yang, Zachary DeVito, Martin Raison, Alykhan
  Tejani, Sasank Chilamkurthy, Benoit Steiner, Lu~Fang, Junjie Bai, and Soumith
  Chintala.
\newblock Pytorch: An imperative style, high-performance deep learning library.
\newblock In \emph{Advances in Neural Information Processing Systems 32}, pages
  8024--8035. Curran Associates, Inc., 2019.

\bibitem[Vaswani et~al.(2017)Vaswani, Shazeer, Parmar, Uszkoreit, Jones, Gomez,
  Kaiser, and Polosukhin]{Transformer}
Ashish Vaswani, Noam Shazeer, Niki Parmar, Jakob Uszkoreit, Llion Jones,
  Aidan~N Gomez, \L~ukasz Kaiser, and Illia Polosukhin.
\newblock Attention is all you need.
\newblock In \emph{Advances in Neural Information Processing Systems},
  volume~30. Curran Associates, Inc., 2017.

\bibitem[Applegate and Cook(1991)]{Orb}
David Applegate and William Cook.
\newblock A computational study of the job-shop scheduling problem.
\newblock \emph{INFORMS Journal on Computing}, 3:\penalty0 149--156, 05 1991.

\bibitem[Kingma and Ba(2014)]{Adam}
Diederik Kingma and Jimmy Ba.
\newblock Adam: A method for stochastic optimization.
\newblock \emph{International Conference on Learning Representations}, 12 2014.

\end{thebibliography}
